\documentclass[11pt,a4paper]{article}
\usepackage[hyperref]{eacl2021}
\usepackage{times}
\usepackage{latexsym}

\usepackage{microtype}

\usepackage{graphicx}
\usepackage{xcolor}

\usepackage{setspace}

\aclfinalcopy % Uncomment this line for the final submission
\title{T-NER: An All-Round Python Library \\ for Transformer-based Named Entity Recognition}

\author{Asahi Ushio \and Jose Camacho-Collados \\
  School of Computer Science and Informatics \\
  Cardiff University, United Kingdom \\
  {\tt \{ushioa,camachocolladosj\}@cardiff.ac.uk}}

\date{}

\begin{document}
\maketitle

%\begin{spacing}{0.99}

\begin{abstract}
Language model (LM) pretraining has led to consistent improvements in many NLP downstream tasks, including named entity recognition (NER). In this paper, we present {\bf T-NER}\footnote{\url{https://github.com/asahi417/tner}} ({\bf T}ransformer-based {\bf N}amed {\bf E}ntity {\bf R}ecognition), a Python library for NER LM finetuning. In addition to its practical utility, T-NER facilitates the study and investigation of the cross-domain and cross-lingual generalization ability of LMs finetuned on NER. Our library also provides a web app where users can get model predictions interactively for arbitrary text, which facilitates qualitative model evaluation for non-expert programmers. We show the potential of the library by compiling nine public NER datasets into a unified format and evaluating the cross-domain and cross-lingual performance across the datasets. The results from our initial experiments show that in-domain performance is generally competitive across datasets. However, cross-domain generalization is challenging even with a large pretrained LM, which has nevertheless capacity to learn domain-specific features if finetuned on a combined dataset. To facilitate future research, we also release all our LM checkpoints via the Hugging Face model hub\footnote{\url{https://huggingface.co/models?search=asahi417/tner}.
}

%including a unified NER model trained on various domains
\end{abstract}

\section{Introduction}

Language model (LM) pretraining has become one of the most common strategies within the natural language processing (NLP) community to solve downstream tasks \citep{peters-etal-2018-deep, howard-ruder-2018-universal, radford2018improving, radford2019language, devlin2018bert}.
LMs trained over large textual data only need to be finetuned on downstream tasks to outperform most of the task-specific designed models. Among the NLP tasks impacted by LM pretraining, named entity recognition (NER) is one of the most prevailing and practical applications. However, the availability of open-source NER libraries for LM training is limited.\footnote{
Recently, spaCy (\url{https://spacy.io/}) has released a general NLP pipeline with pretrained models including a NER feature. Although it provides a very efficient pipeline for processing text, it is not suitable for LM finetuning %as neither finetuning pretrained LMs or 
or %evaluating finetuned LMs 
evaluation on arbitrary NER data.} 
%is supported.}

\begin{figure}[!t]
    \centering
    \includegraphics[width=0.5\textwidth]{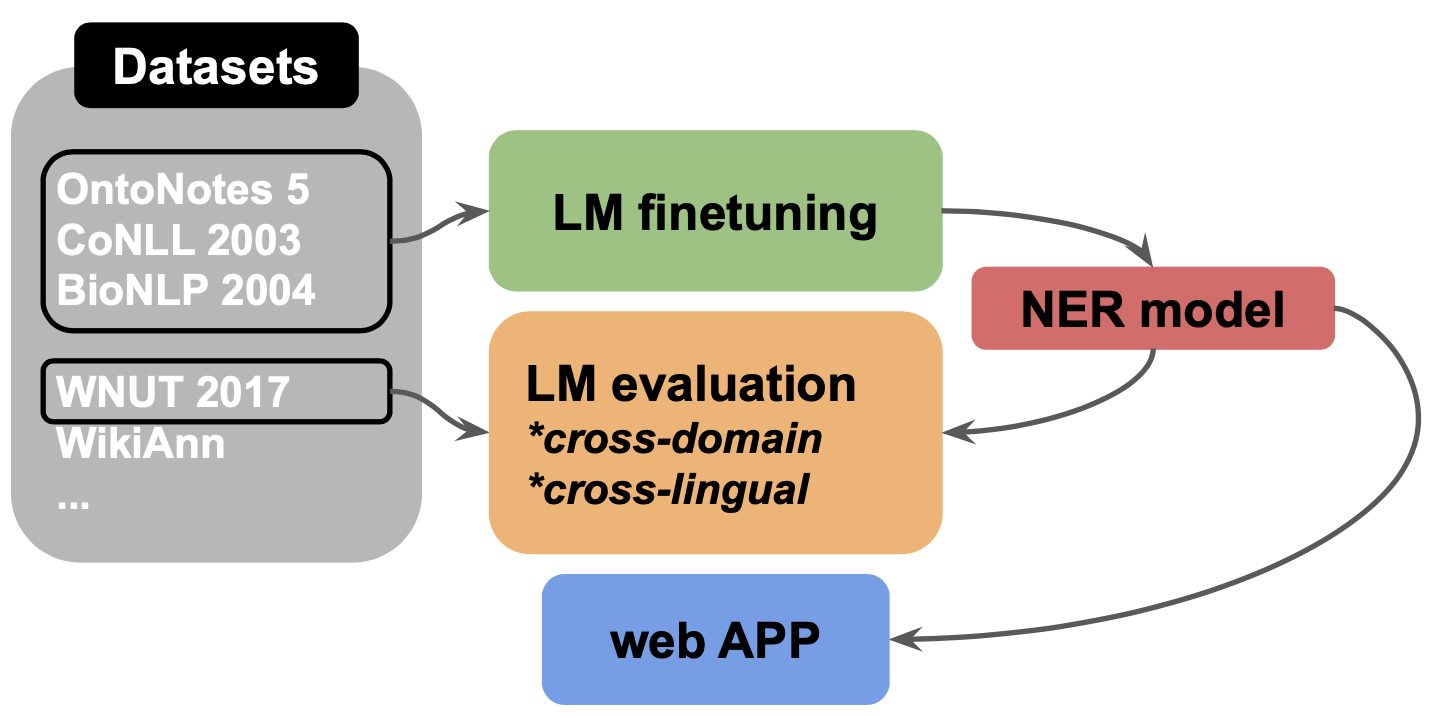}
    \caption{System overview of T-NER.}
    \label{fig:diagram}
\end{figure}

In this paper, we introduce {\bf T-NER}, an open-source Python library for cross-domain analysis for NER with pretrained Transformer-based LMs.
Figure~\ref{fig:diagram} shows a brief overview of our library and its functionalities.
The library facilitates NER experimental design including easy-to-use features such as model training and evaluation. Most notably, it enables to organize cross-domain analyses such as training a NER model and testing it on a different domain, with a small configuration.
We also report initial experiment results, by which we show that although cross-domain NER is challenging, if it has an access to new domains, LM can successfully learn new domain knowledge. The results give us an insight that LM is capable to learn a variety of domain knowledge, but an ordinary finetuning scheme on single dataset most likely causes overfitting and results in poor domain generalization.

As a system design, T-NER is implemented in Pytorch \citep{paszke2019pytorch} on top of the Transformers library \citep{Wolf2019HuggingFacesTS}. Moreover, the interfaces of our training and evaluation modules are highly inspired by Scikit-learn \citep{pedregosa2011scikit}, enabling an interoperability with recent models as well as integrating them in an intuitive way. In addition to the versatility of our toolkit for NER experimentation, we also include an online demo and robust pre-trained models trained across domains.
In the following sections, we provide a brief overview about NER in Section \ref{sec:ner}, explain the system architecture of T-NER with a few basic usages in Section~\ref{sec:system-overview} and describe experiment results on cross-domain transfer with our library in Section~\ref{sec:experiments}.

\begin{figure*}[ht]
    \centering
\includegraphics[width=1\textwidth]{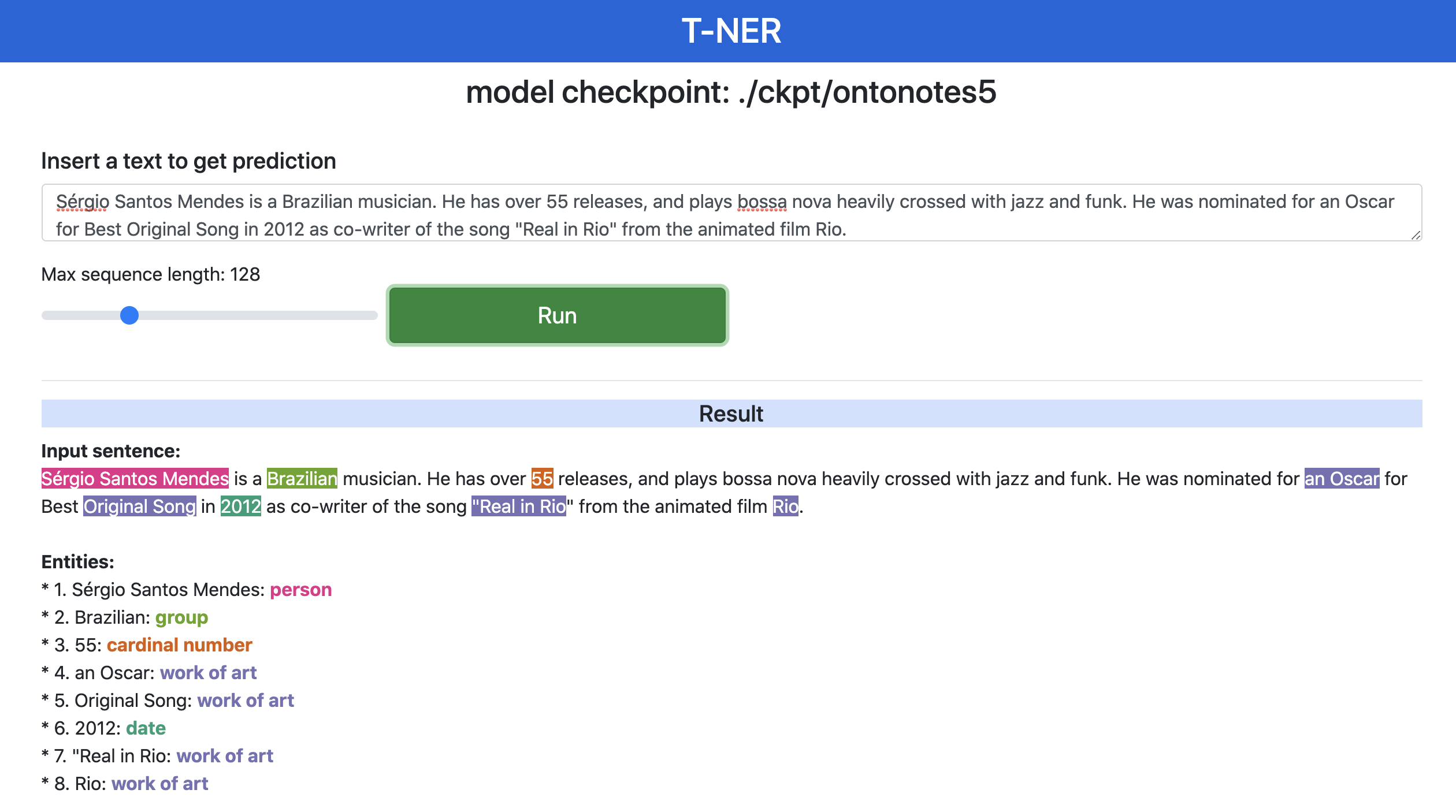}
    \caption{A screenshot from the demo web app. In this example, the NER transformer model is fine-tuned on OntoNotes 5 and a sample sentence is fetched from Wikipedia (\url{en.wikipedia.org/wiki/Sergio_Mendes}).}
    \label{fig:web_app}
\end{figure*}

\section{Named Entity Recognition}
\label{sec:ner}

%Among NLP tasks, named entity recognition (NER) is known as one of the most prevailing and practical applications, where the goal is to detect named entities and identify the entity type given an arbitrary text \citep{tjong-kim-sang-de-meulder-2003-introduction}.
Given an arbitrary text, the task of NER consists of detecting named entities and identifying their type.
For example, given a sentence {\it "Dante was born in Florence."}, a NER model are would identify {\it "Dante"} as a person and {\it "Florence"} as a location. 
Traditionally, NER systems have relied on a classification model on top of hand-engineered feature sets extracted from corpora \citep{ratinov-roth-2009-design, collobert2011natural}, which was improved by carefully designed neural network approaches \citep{lample-etal-2016-neural, chiu-nichols-2016-named, ma-hovy-2016-end}. This paradigm shift was mainly due to its efficient access to contextual information and flexibility, as human-crafted feature sets were no longer required.
Later, contextual representations produced by pretrained LMs have improved the generalization abilities of neural network architectures in many NLP tasks, including NER \citep{peters-etal-2018-deep, devlin2018bert}. %Recently, \newcite{yamada-etal-2020-luke} showed that an entity-aware pretraining objective can enhance LMs by capturing the semantics of entities in an efficient way.  

%Indeed, a key aspect in machine learning theory is to assess the domain generalization capabilities of supervised models.
%Generally, supervised models learnt solely on a training dataset from scratch are prone to fail in unseen domains. Domain adaptation has been proposed to compensate the lack of knowledge of supervised models in new domains \citep{blitzer-etal-2007-biographies, chen2012marginalized, ganin2016domain, bhushan2018deepjdot}.
In particular, LMs \textit{see} millions of plain texts during pretraining, a knowledge that then can be leveraged in downstream NLP applications. This property has been studied in the recently literature by probing their generalization capacity \citep{hendrycks2020pretrained,aharoni-goldberg-2020-unsupervised, desai2020calibration, gururangan2020don}. When it comes to LM generalization studies in NER, the literature is more limited and mainly restricted to in-domain \cite{agarwal2021entity} or multilingual settings \citep{pfeiffer2020mad, hu2020xtreme}. Our library facilitates future research in cross-domain and cross-lingual generalization by providing a unified benchmark for several languages and domain as well as a straightforward implementation of NER LM finetuning.

\section{T-NER: An Overview}
\label{sec:system-overview}

A key design goal was to create a self-contained universal system to train, evaluate, and utilize NER models in an easy way, not only for research purpose but also practical use cases in industry. Moreover, we provide a demo web app (Figure~\ref{fig:web_app}) where users can get predictions from a trained model given a sentence interactively. This way, users (even those without programming experience) can conduct qualitative analyses on their own or existing pre-trained models.

In the following we provide details on the technicalities of the package provided, including details on how to train and evaluate any LM-based architecture. Our package, T-NER, allows practitioners in NLP to get started working on NER with a few lines of code while diving into the recent progress in LM finetuning. We employ Python as our core implementation, as is one of the most prevailing languages in the machine learning and NLP communities. Our library enables Python users to access its various kinds of features such as model training, in- and cross-domain model evaluation, and an interface to get predictions from trained models with minimum effort.

\subsection{Datasets}
\label{ssec:dataset}

For model training and evaluation, we compiled nine public NER datasets from different domains, unifying them into same format:
OntoNotes5 \citep{hovy-etal-2006-ontonotes},
CoNLL 2003 \citep{tjong-kim-sang-de-meulder-2003-introduction},
WNUT 2017 \citep{derczynski-etal-2017-results},
WikiAnn \citep{pan-etal-2017-cross},
FIN \citep{salinas-alvarado-etal-2015-domain},
BioNLP 2004 \citep{collier-kim-2004-introduction},
BioCreative V CDR\footnote{The original dataset consists of long documents which cannot be fed on LM because of the length, so we split them into sentences to reduce their size.} \citep{wei2015overview},
MIT movie review semantic corpus,\footnote{
The movie corpus includes two datasets ({\it eng} and {\it trivia10k13}) coming from different data sources. While both have been integrated into our library, we only used the largest {\it trivia10k13} in our experiments. 
% \red{TOCLARIFY: Any quick reason why using this version and not the other? And are these versions  of the same dataset or actually the same data?}
}
and MIT restaurant review.\footnote{
The original MIT NER corpora can be downloaded from \url{https://groups.csail.mit.edu/sls/downloads/}.} These unified datasets are also made available as part of our T-NER library. Except for WikiAnn that contains 282 languages, all the datasets are in English, and only the MIT corpora are lowercased.
As MIT corpora are commonly used for slot filling task in spoken language understanding 
\citep{liu2017multi}, the characteristics of the entities and annotation guidelines are quite different from the other datasets, but we included them for completeness and to analyze the differences across datasets.

\begin{table*}
\centering
\begin{tabular}{llrr}
\hline
\textbf{Name}  & \textbf{Domain}      & \textbf{Entity types} & \textbf{Data size} \\
\hline
OntoNotes5     & News, Blog, Dialogue & 18 & 59,924/8,582/8,262 \\ \hline
CoNLL 2003     & News                 & 4  & 14,041/3,250/3,453 \\ \hline
WNUT 2017      & SNS                  & 6  & 1,000/1,008/1,287 \\ \hline
WikiAnn        & Wikipedia (282 languages)            & 3  & 20,000/10,000/10,000 \\ \hline
FIN            & Finance              & 4  & 1,164/-/303 \\ \hline
BioNLP 2004    & Biochemical          & 5  & 18,546/-/3,856	\\ \hline
BioCreative V  & Biomedical           & 5  & 5,228/5,330/5,865 \\ \hline
MIT Restaurant & Restaurant review    & 8  & 7,660/-/1,521 \\ \hline
MIT Movie      & Movie review         & 12 & 7,816/-/1,953 \\ \hline
\end{tabular}
\caption{\label{dataset-spec}
Overview of the NER datasets used in our evaluation and included in T-NER. Data size is the number of sentence in training/validation/test set.}
\end{table*}

Table~\ref{dataset-spec} shows statistics of each dataset.   
In Section~\ref{sec:experiments}, we train models on each dataset, and assess the in- and cross-domain accuracy over them.

\paragraph{Dataset format and customization.} Users can utilize their own datasets for both model training and evaluation by formatting them into the IOB scheme \citep{tjong-kim-sang-de-meulder-2003-introduction} which we used to unify all datasets. In the IOB format, all data files contain one word per line with empty lines representing sentence boundaries. At the end of each line there is a tag which states whether the current word is inside a named entity or not. The tag also encodes the type of named entity. Here is an example from CoNLL 2003:

\begin{quote}
\small
\begin{verbatim}
EU B-ORG
rejects O
German B-MISC
call O
to O
boycott O
British B-MISC
lamb O
. O
\end{verbatim}
\end{quote}

\subsection{Model Training}
% Training
% - organize checkpoint
% - tensorboard visualization
% - support recent hardware technique multi GPUs and fp16

We provide modules to facilitate LM finetuning on any given NER dataset. % described in Section~\ref{ssec:dataset}. 
Following \citet{devlin2018bert}, we add a linear layer on top of the last embedding layer in each token, and train all weights with cross-entropy loss. The model training component relies on the Huggingface transformers library \citep{Wolf2019HuggingFacesTS}, one of the largest Python frameworks for distributing pretrained LM checkpoint files. Our library is therefore fully compatible with the Transformers framework: once new model was deployed on the Transformer hub, one can immediately try those models out with our library as a NER model.
To reduce computational complexity, in addition to enabling multi-GPU support, we implement mixture precision during model training by using the apex library\footnote{\url{https://github.com/NVIDIA/apex}}.

The instance of model training in a given dataset\footnote{To use custom datasets, the path to a custom dataset folder can simply be included in the dataset argument.} can be used in an intuitive way as displayed below: 
\begin{quote}
\small
\begin{verbatim}
from tner import TrainTransformersNER
model = TrainTransformersNER(
        dataset="ontonotes5",
        transformer="roberta-base")
model.train()
\end{verbatim}
\end{quote}
With this sample code, we would finetune {\it RoBERTa}$_{BASE}$ \citep{liu2019roberta} on the OntoNotes5 dataset. We also provide an easy extension to train on multiple datasets at the same time:
\begin{quote}
\small
\begin{verbatim}
TrainTransformersNER(
    dataset=[
        "ontonotes5", "wnut2017"
        ],
    transformer="roberta-base")
\end{verbatim}
\end{quote}

Once training is completed, checkpoint files with model weights and other statistics are generated. These are automatically organized for each configuration
and can be easily uploaded to the Hugging Face model hub. 
Ready-to-use code samples can be found in our Google Colab notebook\footnote{\url{https://colab.research.google.com/drive/1AlcTbEsp8W11yflT7SyT0L4C4HG6MXYr?usp=sharing}}, and
details for additional options and arguments are included in the github repository.
Finally, our library supports Tensorboard\footnote{\url{www.tensorflow.org/tensorboard}} to visualize learning curves.

\subsection{Model Evaluation}
Once a NER model is trained, users may want to test the models in the same dataset or a different one to assess its general performance across domains. To this end, we implemented flexible evaluation modules to facilitate cross-domain evaluation comparison, which is also aided by the unification of datasets into the same format (see Section \ref{ssec:dataset}) with a unique label reference lookup. 

The basic usage of the evaluation module is described below.

\begin{quote}
\small
\begin{verbatim}
from tner import TrainTransformersNER
model = TrainTransformersNER(
    "path-to-model-checkpoint"
    )
model.test("ontonotes5")
\end{verbatim}
\end{quote}

Here, the model would be tested on OntoNotes5 dataset, and it could be evaluated on any other test set including custom dataset. As with the model training module, we prepared a Google Colab notebook\footnote{\url{https://colab.research.google.com/drive/1jHVGnFN4AU8uS-ozWJIXXe2fV8HUj8NZ?usp=sharing}} for an example use case, and further details can be found in our github repository.

\section{Evaluation}
\label{sec:experiments}

In this section, we assess the reliability of T-NER with experiments in standard NER datasets.

\subsection{Experimental Setting}
\subsubsection{Implementation details}
Through the experiments, we use {\it XLM-R} \citep{liu2019roberta}, which has shown to be one of the most reliable multi-lingual pretrained LMs for discriminative tasks at the moment. In all experiments we make use of the default configuration and hyperpameters of Huggingface's {\it XLM-R} implementation. For WikiAnn/ja (Japanese), we convert the original character-level tokenization into proper morphological chunk by MeCab\footnote{\url{https://pypi.org/project/mecab-python3/}}.

\subsubsection{Evaluation metrics and protocols}

As customary in the NER literature, we report \textit{span micro-F1 score} computed by seqeval\footnote{\url{https://pypi.org/project/seqeval/}}, a Python library to compute metrics for sequence prediction evaluation.
We refer to this F1 score as \textit{type-aware} F1 score to distinguish it from the the type-ignored metric used to assess the cross-domain performance, which we explain below. 

In a cross-domain evaluation setting, the \textit{type-aware} F1 score easily fails to represent the cross-domain performance if the granularity of entity types differ across datasets.
For instance, the MIT restaurant corpus has entities such as {\it amenity} and {\it rating}, while {\it plot} and {\it actor} are entities from the MIT movie corpus.
Thus, we report {\it type-ignored} F1 score for cross-domain analysis. In this {\it type-ignored} evaluation, the entity type from both of predictions and true labels is disregarded, reducing the task into a simpler entity span detection task. 
This evaluation protocol can be customized by the user at test time.

\subsection{Results}
We conduct three experiments on the nine datasets described in Table~\ref{dataset-spec}: (i) in-domain evaluation (Section \ref{indomainresults}), (ii) cross-domain evaluation (Section \ref{crossdomainresults}), and (iii) cross-lingual evaluation (Section \ref{crosslingualresults}). While the first experiment tests our implementation in standard datasets, the second experiment is aimed at investigating the cross-domain performance of transformer-based NER models.
Finally, as a direct extension of our evaluation module, we show the zero-shot cross-lingual performance of NER models on the WikiAnn dataset. 

\begin{table}
\centering
\begin{tabular}{lrrr}
\hline
\textbf{Dataset} & \textit{BASE} & \textit{LARGE} & \textbf{SoTA} \\\hline
OntoNotes5       & 89.0 & 89.1 & 92.1 \\\hline
CoNLL 2003       & 90.8 & 92.9 & 94.3 \\\hline
WNUT 2017        & 52.8 & 58.5 & 50.3 \\\hline
FIN              & 81.3 & 76.4 & 82.7 \\\hline
BioNLP 2004      & 73.4 & 74.3 & 77.4 \\\hline
BioCreative V    & 88.0 & 88.6 & 89.9 \\\hline
MIT Restaurant   & 79.4 & 79.6 & -    \\\hline
MIT Movie        & 69.9 & 71.2 & -    \\\hline
WikiAnn/en       & 82.7 & 84.0 & 84.8 \\\hline
WikiAnn/ja       & 83.8 & 86.5 & 73.3 \\\hline
WikiAnn/ru       & 88.6 & 90.0 & 91.4 \\\hline
WikiAnn/es       & 90.9 & 92.1 & -    \\\hline
WikiAnn/ko       & 87.5 & 89.6 & -    \\\hline
WikiAnn/ar       & 88.9 & 90.3 & -    \\\hline
\end{tabular}
\caption{\label{in-domain-result}
In-domain \textit{type-aware} F1 score
% of
% {\it XLM-R\textsubscript{BASE}} ({\bf Base}) and 
% {\it XLM-R\textsubscript{LARGE}} ({\bf Large})
for test set on each dataset with current SoTA. 
SoTA on each dataset is attained from the result of
{\it BERT-MRC-DSC} \citep{li2019dice} for OntoNotes5,
{\it LUKE} \citep{yamada-etal-2020-luke} for CoNLL 2003,
{\it CrossWeigh} \citep{wang2019crossweigh} for WNUT 2017,
\citep{pfeiffer2020mad} for WikiAnn (en, ja, ru, es, ko, ar),
\citep{salinas-alvarado-etal-2015-domain} for FIN,
\citep{lee2020biobert} for BioNLP 2004,
\citep{nooralahzadeh2019reinforcement} for BioCreative V
and \citep{pfeiffer2020mad} for WikiAnn/en.
}
\end{table}

\begin{table*}[ht]
\centering
\begin{tabular}{l|rrrrrrr|r}
\hline
\textbf{train\textbackslash{}test} & \textbf{ontonotes}  & \textbf{conll}     & \textbf{wnut}     & \textbf{wiki}             & \textbf{bionlp}    & \textbf{bc5cdr} & \textbf{fin}   & \textbf{avg} \\  \hline
\textbf{ontonotes} & \textbf{91.6}      & 65.4              & 53.6             & 47.5                     & 0.0                   & 0.0               & 18.3          & 40.8 \\ \hline
\textbf{conll}     & 62.2               & \textbf{96.0}     & 69.1             & 61.7                      & 0.0                   & 0.0               & 22.7          & 35.1 \\ \hline
\textbf{wnut}      & 41.8               & 85.7               & \textbf{68.3}    & 54.5                     & 0.0                   & 0.0               & 20.0          & 31.7 \\ \hline
\textbf{wiki}      & 32.8               & 73.3              & 53.6             & \textbf{93.4}            & 0.0                   & 0.0               & 12.2          & 29.6 \\ \hline
\textbf{bionlp}    & 0.0                   & 0.0                  & 0.0                 & 0.0                         & \textbf{79.0}      & 0.0               & 0.0              & 8.7 \\ \hline
\textbf{bc5cdr}    & 0.0                   & 0.0                  & 0.0                 & 0.0                         & 0.0                   & \textbf{88.8}  & 0.0              & 9.8 \\ \hline
\textbf{fin}       & 48.2               & 73.2              & 60.9             & 58.9                     & 0.0                   & 0.0               & \textbf{82.0} & 38.1 \\ \hline\hline
\textbf{all}       & 90.9               & 93.8              & 60.9             & 91.3                     & 78.3               & 84.6           & 75.5          & 81.7 \\ \hline
\end{tabular}
\caption{\label{cross-domain-result}
{\it Type-ignored} F1 score in cross-domain setting over non-lower-cased English datasets.
We compute average of accuracy in each test set, named as {\bf avg}. The model trained on all datasets listed here, is shown as {\bf all}.
}
\end{table*}

\subsubsection{In-domain results}
\label{indomainresults}

The main results are displayed in Table~\ref{in-domain-result}, where we report the \textit{type-aware} F1 score from {\it XLM-R\textsubscript{BASE}} and {\it XLM-R\textsubscript{LARGE}} models along with current state-of-the-art (SoTA).
One can confirm that our framework with {\it XLM-R\textsubscript{LARGE}} achieves a comparable SoTA score, even surpassing it in the WNUT 2017 dataset. In general, {\it XLM-R\textsubscript{LARGE}} performs consistently better than {\it XLM-R\textsubscript{BASE}} but, interestingly, the base model performs better than large on the FIN dataset. %\red{This} is presumably because the large model overfits due to the limited training data. 
This can be attributed to the limited training data in this dataset, which may have caused overfitting in the large model.

Generally, it can be expected to get better accuracy with domain-specific or larger language models that can be integrated into our library. Nonetheless, our goal for these experiments were not to achieve SoTA but rather to provide a competitive and easy-to-use framework. In the remaining experiments we report results for {\it XLM-R\textsubscript{LARGE}} only, but the results for {\it XLM-R\textsubscript{BASE}} can be found in the appendix.

\subsubsection{Cross-domain results}
\label{crossdomainresults}
In this section, we show cross-domain evaluation results on the English datasets\footnote{We excluded the MIT datasets in this setting since they are all lowercased.}: OntoNotes5 (ontonotes), CoNLL 2003 (conll), WNUT 2017 (wnut), WikiAnn/en (wiki), BioNLP 2004 (bionlp), and BioCreative V (bc5cdr), FIN (fin). 
We also report the accuracy of the same XLM-R model trained over a combined dataset resulting from concatenation of all the above datasets. 
% We randomly draw samples from the full dataset for model training. \red{TOVERIFY: "randomly draw samples from the full dataset"? Not sure what that means. Didn't you simply use the training sets of each dataset?}

In Table~\ref{cross-domain-result}, we present the {\it type-ignored} F1 results across datasets. Overall cross-domain scores are not as competitive as in-domain results. This gap reveals the difficulty of transferring NER models into different domains, which may also be attributed to different annotation guidelines or data construction procedures across datasets. Especially, training on the bionlp and bc5cdr datasets lead to a null accuracy when they are evaluated on other datasets, as well as others evaluated on them. Those datasets are very domain specific dataset, as they have entities such as {\it DNA}, {\it Protein}, {\it Chemical}, and {\it Disease}, which results in a poor adaptation to other domains. On the other hand, there are datasets that are more easily transferable, such as wnut and conll. The wnut-trained model achieves 85.7 on the conll dataset and, surprisingly, the conll-trained model actually works better than the wnut-trained model when evaluated on the wnut test set.
This could be also attributed to the data size, as wnut only has 1,000 sentences, while conll has 14,041. Nevertheless, the fact that ontonotes has 59,924 sentences but does not perform better than conll on wnut reveals a certain domain similarity between conll and wnut.

Finally, the model trained on the training sets of all datasets achieves a \textit{type-ignored} F1 score close to the in-domain baselines.
This indicates that a LM is capable of learning representations of different domains. Moreover, leveraging domain similarity as explained above can lead to better results as, for example, distant datasets such as bionlp and bc5cdr surely cause performance drops. This is an example of the type of experiments that could be facilited by T-NER, which we leave for future work.

\subsubsection{Cross-lingual results}
\label{crosslingualresults}

\begin{table}
\centering
\begin{tabular}{l|rrrrrr}
\hline
& \multicolumn{6}{c}{\textbf{test}} \\
\textbf{train} & \textbf{en}    & \textbf{ja}    & \textbf{ru}    & \textbf{ko}    & \textbf{es}    & \textbf{ar}    \\ \hline
\textbf{en} & \textbf{84.0} & 46.3          & 73.1          & 58.1          & 71.4          & 53.2          \\\hline
\textbf{ja} & 53.0           & \textbf{86.5} & 45.7          & 57.1           & 74.5          & 55.4          \\\hline
\textbf{ru} & 60.4          & 53.3          & \textbf{90.0} & 68.1          & 76.8          & 54.9          \\\hline
\textbf{ko} & 57.8          & 62.0          & 68.6          & \textbf{89.6} & 66.2          & 57.2           \\\hline
\textbf{es} & 70.5          & 50.6          & 75.8          & 61.8          & \textbf{92.1} & 62.1          \\\hline
\textbf{ar} & 60.1          & 55.7          & 55.7          & 70.7          & 79.7          & \textbf{90.3} \\ \hline
\end{tabular}
\caption{\label{cross-lingual-result}
Cross-lingual \textit{type-aware} F1 results on various languages for the WikiAnn dataset.}

\end{table}

Finally, we present some results for zero-shot cross-lingual NER over the WikiAnn dataset, where we include six distinct languages: English (en), Japanese (ja), Russian (ru), Korean (ko), Spanish (es), and Arabic (ar). In Table~\ref{cross-lingual-result}, we show the cross-lingual evaluation results. The diagonal includes the results of the model trained on the training data of the same target language. 
There are a few interesting findings. First, we observe a high correlation between Russian and Spanish, which are generally considered to be distant languages and do not share the alphabet.
Second, Arabic also transfers well to Spanish which, despite the Arabic (lexical) influence on the Spanish language \cite{stewart1999spanish}, are still languages from distant families.

Clearly, this is a shallow cross-lingual analysis, but it highlights the possibilities of our library for research in cross-lingual NER. 
Recently, \citep{hu20icml} proposed a compilation of multilingual benchmark tasks including the WikiAnn datasets as a part of it, and {\it XLM-R} proved to be a strong baseline on multilingual NER. This is in line with the results of \newcite{conneau-etal-2020-emerging}, which showed a high capacity of zero-shot cross-lingual transferability. On this respect, \newcite{pfeiffer-etal-2020-mad} proposed a language/task specific adapter module that can further improve cross-lingual adaptation in NER. 
Given the possibilities and recent advances in cross-lingual language models in recent years, we expect our library to help practitioners to experiment and test these advances in NER.

\section{Conclusion}
In this paper, we have presented a Python library to get started with Transformer-based NER models. This paper especially focuses on LM finetuning, and empirically shows the difficulty of cross-domain generalization in NER. 
Our framework is designed to be as simple as possible so that any level of users can start running experiments on NER on any given dataset. To this end, we have also facilitated the evaluation by unifying some of the most popular NER datasets in the literature, including languages other than English. We believe that our initial experiment results emphasize the importance of NER generalization analysis, for which we hope that our open-source library can help NLP community to convey relevant research in an efficient and accessible way.

\section*{Acknowledgements}
We would like to thank Dimosthenis Antypas for testing our library and the anonymous reviewers for their useful comments.

\bibliography{anthology,eacl2021}

\begin{thebibliography}{38}
\expandafter\ifx\csname natexlab\endcsname\relax\def\natexlab#1{#1}\fi

\bibitem[{Agarwal et~al.(2021)Agarwal, Yang, Wallace, and
  Nenkova}]{agarwal2021entity}
Oshin Agarwal, Yinfei Yang, Byron~C Wallace, and Ani Nenkova. 2021.
\newblock Entity-switched datasets: An approach to auditing the in-domain
  robustness of named entity recognition models.
\newblock \emph{arXiv preprint arXiv:2004.04123}.

\bibitem[{Aharoni and Goldberg(2020)}]{aharoni-goldberg-2020-unsupervised}
Roee Aharoni and Yoav Goldberg. 2020.
\newblock \href {https://doi.org/10.18653/v1/2020.acl-main.692} {Unsupervised
  domain clusters in pretrained language models}.
\newblock In \emph{Proceedings of the 58th Annual Meeting of the Association
  for Computational Linguistics}, pages 7747--7763, Online. Association for
  Computational Linguistics.

\bibitem[{Chiu and Nichols(2016)}]{chiu-nichols-2016-named}
Jason~P.C. Chiu and Eric Nichols. 2016.
\newblock \href {https://doi.org/10.1162/tacl_a_00104} {Named entity
  recognition with bidirectional {LSTM}-{CNN}s}.
\newblock \emph{Transactions of the Association for Computational Linguistics},
  4:357--370.

\bibitem[{Collier and Kim(2004)}]{collier-kim-2004-introduction}
Nigel Collier and Jin-Dong Kim. 2004.
\newblock \href {https://www.aclweb.org/anthology/W04-1213} {Introduction to
  the bio-entity recognition task at {JNLPBA}}.
\newblock In \emph{Proceedings of the International Joint Workshop on Natural
  Language Processing in Biomedicine and its Applications
  ({NLPBA}/{B}io{NLP})}, pages 73--78, Geneva, Switzerland. COLING.

\bibitem[{Collobert et~al.(2011)Collobert, Weston, Bottou, Karlen, Kavukcuoglu,
  and Kuksa}]{collobert2011natural}
Ronan Collobert, Jason Weston, L{\'e}on Bottou, Michael Karlen, Koray
  Kavukcuoglu, and Pavel Kuksa. 2011.
\newblock Natural language processing (almost) from scratch.
\newblock \emph{Journal of machine learning research}, 12(ARTICLE):2493--2537.

\bibitem[{Conneau et~al.(2020)Conneau, Wu, Li, Zettlemoyer, and
  Stoyanov}]{conneau-etal-2020-emerging}
Alexis Conneau, Shijie Wu, Haoran Li, Luke Zettlemoyer, and Veselin Stoyanov.
  2020.
\newblock \href {https://doi.org/10.18653/v1/2020.acl-main.536} {Emerging
  cross-lingual structure in pretrained language models}.
\newblock In \emph{Proceedings of the 58th Annual Meeting of the Association
  for Computational Linguistics}, pages 6022--6034, Online. Association for
  Computational Linguistics.

\bibitem[{Derczynski et~al.(2017)Derczynski, Nichols, van Erp, and
  Limsopatham}]{derczynski-etal-2017-results}
Leon Derczynski, Eric Nichols, Marieke van Erp, and Nut Limsopatham. 2017.
\newblock \href {https://doi.org/10.18653/v1/W17-4418} {Results of the
  {WNUT}2017 shared task on novel and emerging entity recognition}.
\newblock In \emph{Proceedings of the 3rd Workshop on Noisy User-generated
  Text}, pages 140--147, Copenhagen, Denmark. Association for Computational
  Linguistics.

\bibitem[{Desai and Durrett(2020)}]{desai2020calibration}
Shrey Desai and Greg Durrett. 2020.
\newblock Calibration of pre-trained transformers.
\newblock \emph{arXiv preprint arXiv:2003.07892}.

\bibitem[{Devlin et~al.(2019)Devlin, Chang, Lee, and
  Toutanova}]{devlin2018bert}
Jacob Devlin, Ming-Wei Chang, Kenton Lee, and Kristina Toutanova. 2019.
\newblock \href {https://doi.org/10.18653/v1/N19-1423} {{BERT}: Pre-training of
  deep bidirectional transformers for language understanding}.
\newblock In \emph{Proceedings of the 2019 Conference of the North {A}merican
  Chapter of the Association for Computational Linguistics: Human Language
  Technologies, Volume 1 (Long and Short Papers)}, pages 4171--4186,
  Minneapolis, Minnesota. Association for Computational Linguistics.

\bibitem[{Gururangan et~al.(2020)Gururangan, Marasovi{\'c}, Swayamdipta, Lo,
  Beltagy, Downey, and Smith}]{gururangan2020don}
Suchin Gururangan, Ana Marasovi{\'c}, Swabha Swayamdipta, Kyle Lo, Iz~Beltagy,
  Doug Downey, and Noah~A Smith. 2020.
\newblock Don't stop pretraining: Adapt language models to domains and tasks.
\newblock \emph{arXiv preprint arXiv:2004.10964}.

\bibitem[{Hendrycks et~al.(2020)Hendrycks, Liu, Wallace, Dziedzic, Krishnan,
  and Song}]{hendrycks2020pretrained}
Dan Hendrycks, Xiaoyuan Liu, Eric Wallace, Adam Dziedzic, Rishabh Krishnan, and
  Dawn Song. 2020.
\newblock Pretrained transformers improve out-of-distribution robustness.
\newblock \emph{arXiv preprint arXiv:2004.06100}.

\bibitem[{Hovy et~al.(2006)Hovy, Marcus, Palmer, Ramshaw, and
  Weischedel}]{hovy-etal-2006-ontonotes}
Eduard Hovy, Mitchell Marcus, Martha Palmer, Lance Ramshaw, and Ralph
  Weischedel. 2006.
\newblock \href {https://www.aclweb.org/anthology/N06-2015} {{O}nto{N}otes: The
  90{\%} solution}.
\newblock In \emph{Proceedings of the Human Language Technology Conference of
  the {NAACL}, Companion Volume: Short Papers}, pages 57--60, New York City,
  USA. Association for Computational Linguistics.

\bibitem[{Howard and Ruder(2018)}]{howard-ruder-2018-universal}
Jeremy Howard and Sebastian Ruder. 2018.
\newblock \href {https://doi.org/10.18653/v1/P18-1031} {Universal language
  model fine-tuning for text classification}.
\newblock In \emph{Proceedings of the 56th Annual Meeting of the Association
  for Computational Linguistics (Volume 1: Long Papers)}, pages 328--339,
  Melbourne, Australia. Association for Computational Linguistics.

\bibitem[{Hu et~al.(2020{\natexlab{a}})Hu, Ruder, Siddhant, Neubig, Firat, and
  Johnson}]{hu20icml}
Junjie Hu, Sebastian Ruder, Aditya Siddhant, Graham Neubig, Orhan Firat, and
  Melvin Johnson. 2020{\natexlab{a}}.
\newblock \href {https://arxiv.org/pdf/2003.11080.pdf} {{XTREME}: A massively
  multilingual multi-task benchmark for evaluating cross-lingual
  generalisation}.
\newblock In \emph{International Conference on Machine Learning (ICML)}.

\bibitem[{Hu et~al.(2020{\natexlab{b}})Hu, Ruder, Siddhant, Neubig, Firat, and
  Johnson}]{hu2020xtreme}
Junjie Hu, Sebastian Ruder, Aditya Siddhant, Graham Neubig, Orhan Firat, and
  Melvin Johnson. 2020{\natexlab{b}}.
\newblock Xtreme: A massively multilingual multi-task benchmark for evaluating
  cross-lingual generalization.
\newblock \emph{arXiv preprint arXiv:2003.11080}.

\bibitem[{Lample et~al.(2016)Lample, Ballesteros, Subramanian, Kawakami, and
  Dyer}]{lample-etal-2016-neural}
Guillaume Lample, Miguel Ballesteros, Sandeep Subramanian, Kazuya Kawakami, and
  Chris Dyer. 2016.
\newblock \href {https://doi.org/10.18653/v1/N16-1030} {Neural architectures
  for named entity recognition}.
\newblock In \emph{Proceedings of the 2016 Conference of the North {A}merican
  Chapter of the Association for Computational Linguistics: Human Language
  Technologies}, pages 260--270, San Diego, California. Association for
  Computational Linguistics.

\bibitem[{Lee et~al.(2020)Lee, Yoon, Kim, Kim, Kim, So, and
  Kang}]{lee2020biobert}
Jinhyuk Lee, Wonjin Yoon, Sungdong Kim, Donghyeon Kim, Sunkyu Kim, Chan~Ho So,
  and Jaewoo Kang. 2020.
\newblock Biobert: a pre-trained biomedical language representation model for
  biomedical text mining.
\newblock \emph{Bioinformatics}, 36(4):1234--1240.

\bibitem[{Li et~al.(2019)Li, Sun, Meng, Liang, Wu, and Li}]{li2019dice}
Xiaoya Li, Xiaofei Sun, Yuxian Meng, Junjun Liang, Fei Wu, and Jiwei Li. 2019.
\newblock Dice loss for data-imbalanced nlp tasks.
\newblock \emph{arXiv preprint arXiv:1911.02855}.

\bibitem[{Liu and Lane(2017)}]{liu2017multi}
Bing Liu and Ian Lane. 2017.
\newblock Multi-domain adversarial learning for slot filling in spoken language
  understanding.
\newblock \emph{arXiv preprint arXiv:1711.11310}.

\bibitem[{Liu et~al.(2019)Liu, Ott, Goyal, Du, Joshi, Chen, Levy, Lewis,
  Zettlemoyer, and Stoyanov}]{liu2019roberta}
Yinhan Liu, Myle Ott, Naman Goyal, Jingfei Du, Mandar Joshi, Danqi Chen, Omer
  Levy, Mike Lewis, Luke Zettlemoyer, and Veselin Stoyanov. 2019.
\newblock Roberta: A robustly optimized bert pretraining approach.
\newblock \emph{arXiv preprint arXiv:1907.11692}.

\bibitem[{Ma and Hovy(2016)}]{ma-hovy-2016-end}
Xuezhe Ma and Eduard Hovy. 2016.
\newblock \href {https://doi.org/10.18653/v1/P16-1101} {End-to-end sequence
  labeling via bi-directional {LSTM}-{CNN}s-{CRF}}.
\newblock In \emph{Proceedings of the 54th Annual Meeting of the Association
  for Computational Linguistics (Volume 1: Long Papers)}, pages 1064--1074,
  Berlin, Germany. Association for Computational Linguistics.

\bibitem[{Nooralahzadeh et~al.(2019)Nooralahzadeh, L{\o}nning, and
  {\O}vrelid}]{nooralahzadeh2019reinforcement}
Farhad Nooralahzadeh, Jan~Tore L{\o}nning, and Lilja {\O}vrelid. 2019.
\newblock Reinforcement-based denoising of distantly supervised ner with
  partial annotation.
\newblock In \emph{Proceedings of the 2nd Workshop on Deep Learning Approaches
  for Low-Resource NLP (DeepLo 2019)}, pages 225--233.

\bibitem[{Pan et~al.(2017)Pan, Zhang, May, Nothman, Knight, and
  Ji}]{pan-etal-2017-cross}
Xiaoman Pan, Boliang Zhang, Jonathan May, Joel Nothman, Kevin Knight, and Heng
  Ji. 2017.
\newblock \href {https://doi.org/10.18653/v1/P17-1178} {Cross-lingual name
  tagging and linking for 282 languages}.
\newblock In \emph{Proceedings of the 55th Annual Meeting of the Association
  for Computational Linguistics (Volume 1: Long Papers)}, pages 1946--1958,
  Vancouver, Canada. Association for Computational Linguistics.

\bibitem[{Paszke et~al.(2019)Paszke, Gross, Massa, Lerer, Bradbury, Chanan,
  Killeen, Lin, Gimelshein, Antiga et~al.}]{paszke2019pytorch}
Adam Paszke, Sam Gross, Francisco Massa, Adam Lerer, James Bradbury, Gregory
  Chanan, Trevor Killeen, Zeming Lin, Natalia Gimelshein, Luca Antiga, et~al.
  2019.
\newblock Pytorch: An imperative style, high-performance deep learning library.
\newblock In \emph{Advances in neural information processing systems}, pages
  8026--8037.

\bibitem[{Pedregosa et~al.(2011)Pedregosa, Varoquaux, Gramfort, Michel,
  Thirion, Grisel, Blondel, Prettenhofer, Weiss, Dubourg
  et~al.}]{pedregosa2011scikit}
Fabian Pedregosa, Ga{\"e}l Varoquaux, Alexandre Gramfort, Vincent Michel,
  Bertrand Thirion, Olivier Grisel, Mathieu Blondel, Peter Prettenhofer, Ron
  Weiss, Vincent Dubourg, et~al. 2011.
\newblock Scikit-learn: Machine learning in python.
\newblock \emph{the Journal of machine Learning research}, 12:2825--2830.

\bibitem[{Peters et~al.(2018)Peters, Neumann, Iyyer, Gardner, Clark, Lee, and
  Zettlemoyer}]{peters-etal-2018-deep}
Matthew Peters, Mark Neumann, Mohit Iyyer, Matt Gardner, Christopher Clark,
  Kenton Lee, and Luke Zettlemoyer. 2018.
\newblock \href {https://doi.org/10.18653/v1/N18-1202} {Deep contextualized
  word representations}.
\newblock In \emph{Proceedings of the 2018 Conference of the North {A}merican
  Chapter of the Association for Computational Linguistics: Human Language
  Technologies, Volume 1 (Long Papers)}, pages 2227--2237, New Orleans,
  Louisiana. Association for Computational Linguistics.

\bibitem[{Pfeiffer et~al.(2020{\natexlab{a}})Pfeiffer, Vuli{\'c}, Gurevych, and
  Ruder}]{pfeiffer2020mad}
Jonas Pfeiffer, Ivan Vuli{\'c}, Iryna Gurevych, and Sebastian Ruder.
  2020{\natexlab{a}}.
\newblock Mad-x: An adapter-based framework for multi-task cross-lingual
  transfer.
\newblock \emph{arXiv preprint arXiv:2005.00052}.

\bibitem[{Pfeiffer et~al.(2020{\natexlab{b}})Pfeiffer, Vuli{\'c}, Gurevych, and
  Ruder}]{pfeiffer-etal-2020-mad}
Jonas Pfeiffer, Ivan Vuli{\'c}, Iryna Gurevych, and Sebastian Ruder.
  2020{\natexlab{b}}.
\newblock \href {https://doi.org/10.18653/v1/2020.emnlp-main.617} {{MAD-X}:
  {A}n {A}dapter-{B}ased {F}ramework for {M}ulti-{T}ask {C}ross-{L}ingual
  {T}ransfer}.
\newblock In \emph{Proceedings of the 2020 Conference on Empirical Methods in
  Natural Language Processing (EMNLP)}, pages 7654--7673, Online. Association
  for Computational Linguistics.

\bibitem[{Radford et~al.(2018)Radford, Narasimhan, Salimans, and
  Sutskever}]{radford2018improving}
Alec Radford, Karthik Narasimhan, Tim Salimans, and Ilya Sutskever. 2018.
\newblock Improving language understanding by generative pre-training.

\bibitem[{Radford et~al.(2019)Radford, Wu, Child, Luan, Amodei, and
  Sutskever}]{radford2019language}
Alec Radford, Jeffrey Wu, Rewon Child, David Luan, Dario Amodei, and Ilya
  Sutskever. 2019.
\newblock Language models are unsupervised multitask learners.

\bibitem[{Ratinov and Roth(2009)}]{ratinov-roth-2009-design}
Lev Ratinov and Dan Roth. 2009.
\newblock \href {https://www.aclweb.org/anthology/W09-1119} {Design challenges
  and misconceptions in named entity recognition}.
\newblock In \emph{Proceedings of the Thirteenth Conference on Computational
  Natural Language Learning ({C}o{NLL}-2009)}, pages 147--155, Boulder,
  Colorado. Association for Computational Linguistics.

\bibitem[{Salinas~Alvarado et~al.(2015)Salinas~Alvarado, Verspoor, and
  Baldwin}]{salinas-alvarado-etal-2015-domain}
Julio~Cesar Salinas~Alvarado, Karin Verspoor, and Timothy Baldwin. 2015.
\newblock \href {https://www.aclweb.org/anthology/U15-1010} {Domain adaption of
  named entity recognition to support credit risk assessment}.
\newblock In \emph{Proceedings of the Australasian Language Technology
  Association Workshop 2015}, pages 84--90, Parramatta, Australia.

\bibitem[{Stewart et~al.(1999)}]{stewart1999spanish}
Miranda Stewart et~al. 1999.
\newblock \emph{The Spanish language today}.
\newblock Psychology Press.

\bibitem[{Tjong Kim~Sang and
  De~Meulder(2003)}]{tjong-kim-sang-de-meulder-2003-introduction}
Erik~F. Tjong Kim~Sang and Fien De~Meulder. 2003.
\newblock \href {https://www.aclweb.org/anthology/W03-0419} {Introduction to
  the {C}o{NLL}-2003 shared task: Language-independent named entity
  recognition}.
\newblock In \emph{Proceedings of the Seventh Conference on Natural Language
  Learning at {HLT}-{NAACL} 2003}, pages 142--147.

\bibitem[{Wang et~al.(2019)Wang, Shang, Liu, Lu, Liu, and
  Han}]{wang2019crossweigh}
Zihan Wang, Jingbo Shang, Liyuan Liu, Lihao Lu, Jiacheng Liu, and Jiawei Han.
  2019.
\newblock Crossweigh: Training named entity tagger from imperfect annotations.
\newblock In \emph{Proceedings of the 2019 Conference on Empirical Methods in
  Natural Language Processing and the 9th International Joint Conference on
  Natural Language Processing (EMNLP-IJCNLP)}, pages 5157--5166.

\bibitem[{Wei et~al.(2015)Wei, Peng, Leaman, Davis, Mattingly, Li, Wiegers, and
  Lu}]{wei2015overview}
Chih-Hsuan Wei, Yifan Peng, Robert Leaman, Allan~Peter Davis, Carolyn~J
  Mattingly, Jiao Li, Thomas~C Wiegers, and Zhiyong Lu. 2015.
\newblock Overview of the biocreative v chemical disease relation (cdr) task.
\newblock In \emph{Proceedings of the fifth BioCreative challenge evaluation
  workshop}, volume~14.

\bibitem[{Wolf et~al.(2019)Wolf, Debut, Sanh, Chaumond, Delangue, Moi, Cistac,
  Rault, Louf, Funtowicz, Davison, Shleifer, von Platen, Ma, Jernite, Plu, Xu,
  Scao, Gugger, Drame, Lhoest, and Rush}]{Wolf2019HuggingFacesTS}
Thomas Wolf, Lysandre Debut, Victor Sanh, Julien Chaumond, Clement Delangue,
  Anthony Moi, Pierric Cistac, Tim Rault, Rémi Louf, Morgan Funtowicz, Joe
  Davison, Sam Shleifer, Patrick von Platen, Clara Ma, Yacine Jernite, Julien
  Plu, Canwen Xu, Teven~Le Scao, Sylvain Gugger, Mariama Drame, Quentin Lhoest,
  and Alexander~M. Rush. 2019.
\newblock Huggingface's transformers: State-of-the-art natural language
  processing.
\newblock \emph{ArXiv}, abs/1910.03771.

\bibitem[{Yamada et~al.(2020)Yamada, Asai, Shindo, Takeda, and
  Matsumoto}]{yamada-etal-2020-luke}
Ikuya Yamada, Akari Asai, Hiroyuki Shindo, Hideaki Takeda, and Yuji Matsumoto.
  2020.
\newblock \href {https://doi.org/10.18653/v1/2020.emnlp-main.523} {{LUKE}: Deep
  contextualized entity representations with entity-aware self-attention}.
\newblock In \emph{Proceedings of the 2020 Conference on Empirical Methods in
  Natural Language Processing (EMNLP)}, pages 6442--6454, Online. Association
  for Computational Linguistics.

\end{thebibliography}
\bibliographystyle{acl_natbib}

\appendix

\section{Appendices}
\label{sec:appendix}

In all experiments we make use of the default configuration and hyperpameters of Huggingface's {\it XLM-R} implementation.

\subsection{Cross-lingual Results}
In this section, we show cross-lingual analysis on {\it XLM-R\textsubscript{BASE}}, where the result is shown in Table~\ref{app:cross-lingual-base}. For these cross-lingual results, we rely on the WikiAnn dataset where zero-shot cross-lingual NER over six distinct languages is conducted: English (en), Japanese (ja), Russian (ru), Korean (ko), Spanish (es), and Arabic (ar).

\subsection{Cross-domain Results}
In this section, we show a few more results on our cross-domain analysis, which is based on non-lowercased English datasets: OntoNotes5 (ontonotes), CoNLL 2003 (conll), WNUT 2017 (wnut), WikiAnn/en (wiki), BioNLP 2004 (bionlp), and BioCreative V (bc5cdr), and FIN (fin).
Table~\ref{app:cross-domain-all} shows the type-aware F1 score of the {\it XLM-R\textsubscript{LARGE}} and {\it XLM-R\textsubscript{BASE}} models trained on all the datasets. Furthermore, Table~\ref{app:cross-domain-base} shows additional results for {\it XLM-R\textsubscript{BASE}} in the type-ignored evaluation.

\begin{table}[ht]
\centering
\begin{tabular}{l|rrrrrrr}
\hline
& \multicolumn{6}{c}{\textbf{test}} \\
\textbf{train} & \textbf{en}   & \textbf{ja}   & \textbf{ru}   & \textbf{ko}   & \textbf{es}   & \textbf{ar}    \\\hline
\textbf{en}    & \textbf{82.8} & 38.6          & 65.7          & 50.4          & 73.8          & 44.5           \\\hline
\textbf{ja}    & 53.8          & \textbf{83.9} & 46.9          & 60.1          & 71.3          & 46.3           \\\hline
\textbf{ru}    & 51.9          & 39.9          & \textbf{88.7} & 51.9          & 66.8          & 51.0           \\\hline
\textbf{ko}    & 54.7          & 51.6          & 53.3          & \textbf{87.5} & 63.3          & 52.3           \\\hline
\textbf{es}    & 65.7          & 44.0          & 66.5          & 54.1          & \textbf{90.9} & 59.4           \\\hline
\textbf{ar}    & 53.1          & 49.2          & 49.4          & 59.7          & 73.6          & \textbf{88.9}  \\\hline
\end{tabular}
\caption{\label{app:cross-lingual-base}
Cross-lingual \textit{type-aware} F1 score over WikiAnn dataset with {\it XLM-R\textsubscript{BASE}}.}
\end{table}

\begin{table}[ht]
\begin{tabular}{l|rrrr}
\hline
 &\multicolumn{2}{c}{\textbf{uppercase}}& \multicolumn{2}{c}{\textbf{lowercase}} \\
\textbf{Datasets} & {\it BASE} & {\it LARGE } & \textit{BASE} & \textit{LARGE} \\ \hline
\textbf{ontonotes}                 & 85.8          & 87.8           & 81.7                      & 85.6                       \\\hline
\textbf{conll}                     & 87.2          & 90.3           & 82.8                      & 87.6                       \\\hline
\textbf{wnut}                      & 49.6          & 55.1           & 43.7                      & 51.3                       \\\hline
\textbf{wiki}                      & 79.1          & 82.7           & 75.2                      & 80.8                       \\\hline
\textbf{bionlp}                    & 72.9          & 74.1           & 71.7                      & 74.0                       \\\hline
\textbf{bc5cdr}                    & 79.4          & 85.0           & 78.0                      & 84.2                       \\\hline
\textbf{fin}                       & 72.4          & 72.4           & 72.4                      & 73.5                       \\\hline
\textbf{restaurant}                & -             & -              & 76.8                      & 80.9                       \\\hline
\textbf{movie}                     & -             & -              & 67.8                      & 71.8                       \\\hline
\end{tabular}
\caption{\label{app:cross-domain-all}
\textit{Type-aware} F1 score across different test sets of models trained on all {\bf uppercase}/{\bf lowercase} English datasets with {\it XLM-R\textsubscript{BASE}} or {\it XLM-R\textsubscript{LARGE}}.
}
\end{table}

\begin{table*}[ht]
\centering
\begin{tabular}{l|rrrrrrr|r}
\hline
\textbf{train\textbackslash{}test} & \textbf{ontonotes} & \textbf{conll} & \textbf{wnut} & \textbf{wiki} & \textbf{bionlp} & \textbf{bc5cdr} & \textbf{fin}  & \textbf{avg} \\ \hline
\textbf{ontonotes}                & \textbf{91.8}       & 62.2               & 51.7              & 44.7                      & 0.0                 & 0.0             & 31.8          & 40.3         \\ \hline
\textbf{conll}                 & 60.5                & \textbf{95.7}      & 66.6              & 60.8                      & 0.0                 & 0.0             & 33.5          & 45.3         \\ \hline
\textbf{wnut}                  & 41.3                & 81.3               & \textbf{63.0}     & 56.3                      & 0.0                 & 0.0             & 20.5          & 37.5         \\ \hline
\textbf{wiki}          & 30.2                & 71.8               & 45.3              & \textbf{92.6}             & 0.0                 & 0.0             & 11.5          & 35.9         \\ \hline
\textbf{bionlp}                & 0.0                 & 0.0                & 0.0               & 0.0                       & \textbf{78.5}       & 0.0             & 0.0           & 11.2         \\ \hline
\textbf{bc5cdr}                    & 0.0                 & 0.0                & 0.0               & 0.0                       & 0.0                 & \textbf{87.5}   & 0.0           & 12.5         \\ \hline
\textbf{fin}                       & 49.0                & 73.5               & 62.2              & 60.7                      & 0.0                 & 0.0             & \textbf{82.8} & 46.9         \\ \hline \hline
\textbf{all}              & 89.7                & 92.4               & 55.8              & 89.3                      & 78.2                & 80.0            & 74.8          & 80.0        \\ \hline
\end{tabular}
\caption{\label{app:cross-domain-base}
{\it Type-ignored} F1 score in cross-domain setting over non-lower-cased English datasets with {\it XLM-R\textsubscript{BASE}}.
We compute average of accuracy in each test set, named as {\bf avg}. The model trained on all datasets listed here, is shown as {\bf all}.
}
\end{table*}

\paragraph{Cross-domain results with lowercased datasets. }
In this section, we show cross-domain results on the English datasets including lowercased corpora such as MIT Restaurant (restaurant) and MIT Movie (movie). Since those datasets are lowercasd, we converted all datasets into lowercase.
Tables~\ref{app:cross-domain-large-lower} and Table~\ref{app:cross-domain-base-lower} show the {\it type-ignored} F1 score across models trained on different English datasets including lowercased corpora with {\it XLM-R\textsubscript{LARGE}} and {\it XLM-R\textsubscript{BASE}}, respectively.

\begin{table*}[ht]
\centering
\begin{tabular}{l|rrrrrrrrr|r}
\hline
\textbf{train\textbackslash{}test} & \textbf{ontonotes} & \textbf{conll} & \textbf{wnut} & \textbf{wiki} & \textbf{bionlp} & \textbf{bc5cdr} & \textbf{fin} & \textbf{restaurant} & \textbf{movie} & \textbf{avg} \\\hline
\textbf{ontonotes}  & 89.3               & 59.9           & 50.1          & 44.7          & 0.0               & 0.0               & 15.1         & 4.5                 & 88.6           & 39.1         \\\hline
\textbf{conll}      & 57.7               & 94.8           & 67.0          & 57.9          & 0.0               & 0.0               & 20.5         & 23.9                & 0.0              & 35.7         \\\hline
\textbf{wnut}       & 39.8               & 80.3           & 61.3          & 52.3          & 0.0               & 0.0               & 19.5         & 18.8                & 0.0              & 30.2         \\\hline
\textbf{wiki}       & 28.5               & 69.7           & 51.2          & 92.4          & 0.0               & 0.0               & 12.0           & 3.0                 & 0.0              & 28.5         \\\hline
\textbf{bionlp}     & 0.0                  & 0.0              & 0.0             & 0.0             & 79.0          & 0.0               & 0.0            & 0.0                   & 0.0              & 8.7          \\\hline
\textbf{bc5cdr}     & 0.0                  & 0.0              & 0.0             & 0.0             & 0.0               & 88.9            & 0.0            & 0.0                   & 0.0              & 9.8          \\\hline
\textbf{fin}        & 46                 & 72.0             & 61.5          & 54.8          & 0.0               & 0.0               & 83.0           & 24.5                & 0.0              & 37.9         \\\hline
\textbf{restaurant} & 4.6                & 21.7           & 22.9          & 22.3          & 0.0               & 0.0               & 5.4          & 83.4                & 0.0              & 17.8         \\\hline
\textbf{movie}      & 10.9               & 0.0              & 0.0             & 0.0             & 0.0               & 0.0               & 0.0            & 0.0                   & 73.1           & 9.3          \\\hline\hline
\textbf{all}        & 88.5               & 92.1           & 58.0            & 90.0            & 79.0              & 84.6            & 74.5         & 85.3                & 74.1           & 80.7        \\\hline
\end{tabular}
\caption{\label{app:cross-domain-large-lower}
\textit{Type-ignored} F1 score in cross-domain setting over lower-cased English datasets with {\it XLM-R\textsubscript{LARGE}}.
We compute average of accuracy in each test set, named as {\bf avg}. The model trained on all datasets listed here, is shown as {\bf all}.}
\end{table*}

\begin{table*}[ht]
\begin{tabular}{l|rrrrrrrrr|r}
\hline
\textbf{train\textbackslash{}test} & \textbf{ontonotes} & \textbf{conll} & \textbf{wnut} & \textbf{wiki} & \textbf{bionlp} & \textbf{bc5cdr} & \textbf{fin}  & \textbf{restaurant} & \textbf{movie} & \textbf{avg} \\ \hline
\textbf{ontonotes}                 & \textbf{88.3}      & 56.7           & 49.0          & 41.4          & 0.0             & 0.0             & 11.7          & 4.2                 & 88.3           & 37.7         \\ \hline
\textbf{conll}                     & 55.1               & \textbf{93.7}  & 60.5          & 56.8          & 0.0             & 0.0             & 20.4          & 21.9                & 0.0            & 34.3         \\\hline
\textbf{wnut}                      & 38.1               & 73.0           & \textbf{57.5} & 49.1          & 0.0             & 0.0             & 21.1          & 20.4                & 0.0            & 28.8         \\\hline
\textbf{wiki}                      & 26.3               & 66.5           & 41.4          & \textbf{90.9} & 0.0             & 0.0             & 9.7           & 7.6                 & 0.0            & 26.9         \\\hline
\textbf{bionlp}                    & 0.0                & 0.0            & 0.0           & 0.0           & \textbf{78.7}   & 0.0             & 0.0           & 0.0                 & 0.0            & 8.7          \\\hline
\textbf{bc5cdr}                    & 0.0                & 0.0            & 0.0           & 0.0           & 0.0             & \textbf{88.0}   & 0.0           & 0.0                 & 0.0            & 9.8          \\\hline
\textbf{fin}                       & 41.3               & 64.4           & 45.8          & 57.8          & 0.0             & 0.0             & \textbf{81.5} & 22.0                & 0.0            & 34.8         \\\hline
\textbf{restaurant}                & 8.1                & 19.1           & 19.6          & 19.1          & 0.0             & 0.0             & 13.5          & \textbf{83.6}       & 0.0            & 18.1         \\\hline
\textbf{movie}                     & 14.5               & 0.0            & 0.0           & 0.0           & 0.0             & 0.0             & 0.0           & 0.0                 & \textbf{73.1}  & 9.7          \\\hline\hline
\textbf{all}                       & 86.1               & 89.5           & 49.9          & 86.2          & 76.9            & 78.8            & 75.4          & 82.4                & 72.2           & 77.5        \\\hline
\end{tabular}
\caption{\label{app:cross-domain-base-lower}
\textit{Type-ignored} F1 score in cross-domain setting over lower-cased English datasets with {\it XLM-R\textsubscript{BASE}}.
We compute average of accuracy in each test set, named as {\bf avg}. The model trained on all datasets listed here, is shown as {\bf all}.}
\end{table*}

%\end{spacing}

\end{document}